\def\year{2020}\relax
\documentclass[letterpaper]{article} 
\usepackage{aaai20}  
\usepackage{times}  
\usepackage{helvet} 
\usepackage{courier}  
\usepackage[hyphens]{url}  
\usepackage{graphicx} 
\usepackage{graphicx}  
\usepackage{multirow}
\usepackage{amsmath}
\usepackage{amssymb}
\usepackage{hyperref}

\title{Dynamic Graph Representation for Occlusion Handling in Biometrics}

\author{Min Ren\textsuperscript{1 2}, Yunlong Wang\textsuperscript{2}, Zhenan Sun\textsuperscript{2}, Tieniu Tan\textsuperscript{2}\\
\textsuperscript{1}University of Chinese Academy of Sciences, \textsuperscript{2}CRIPAC NLPR CASIA, Beijing, P.R. China\\
\{min.ren, yunlong.wang\}@cripac.ia.ac.cn, \{znsun, tnt\}@nlpr.ia.ac.cn
}

\begin{document}

\maketitle

\begin{abstract}
The generalization ability of Convolutional neural networks (CNNs) for biometrics drops greatly due to the adverse effects of various occlusions. To this end, we propose a novel unified framework integrated the merits of both CNNs and graphical models to learn dynamic graph representations for occlusion problems in biometrics, called Dynamic Graph Representation (DGR). Convolutional features onto certain regions are re-crafted by a graph generator to establish the connections among the spatial parts of biometrics and build Feature Graphs based on these node representations. Each node of Feature Graphs corresponds to a specific part of the input image and the edges express the spatial relationships between parts. By analyzing the similarities between the nodes, the framework is able to adaptively remove the nodes representing the occluded parts. During dynamic graph matching, we propose a novel strategy to measure the distances of both nodes and adjacent matrixes. In this way, the proposed method is more convincing than CNNs-based methods because the dynamic graph method implies a more illustrative and reasonable inference of the biometrics decision. Experiments conducted on iris and face demonstrate the superiority of the proposed framework, which boosts the accuracy of occluded biometrics recognition by a large margin comparing with baseline methods. The code is avaliable at \url{https://github.com/RenMin1991/Dyamic\_Graph\_Representation}

\end{abstract}


\section{Introduction}

Deep learning methods have gained great success in recent years, especially in the area of computer vision. Convolutional Neural Networks (CNNs) have been widely applied as a powerful tool for image classification and feature extraction~\cite{L1998Gradient} \cite{Krizhevsky2012ImageNet} \cite{Simonyan2014Very} \cite{Szegedy2015Going} \cite{He2016Deep} \cite{Huang2017Densely} \cite{Hu2018Squeeze}. CNNs are also universally applied to biometrics, including iris recognition~\cite{Liu2016DeepIris} \cite{Zhao2017Towards} \cite{Zhang2018Deep}, face recognition~\cite{Taigman2014DeepFace} \cite{Yi2014Deep} \cite{Schroff2015FaceNet} \cite{Liu2017SphereFace} \cite{Xiang2018A} \cite{Deng2018ArcFace} and so on.

\begin{figure}[t]
\begin{center}
\includegraphics[width=0.75\linewidth]{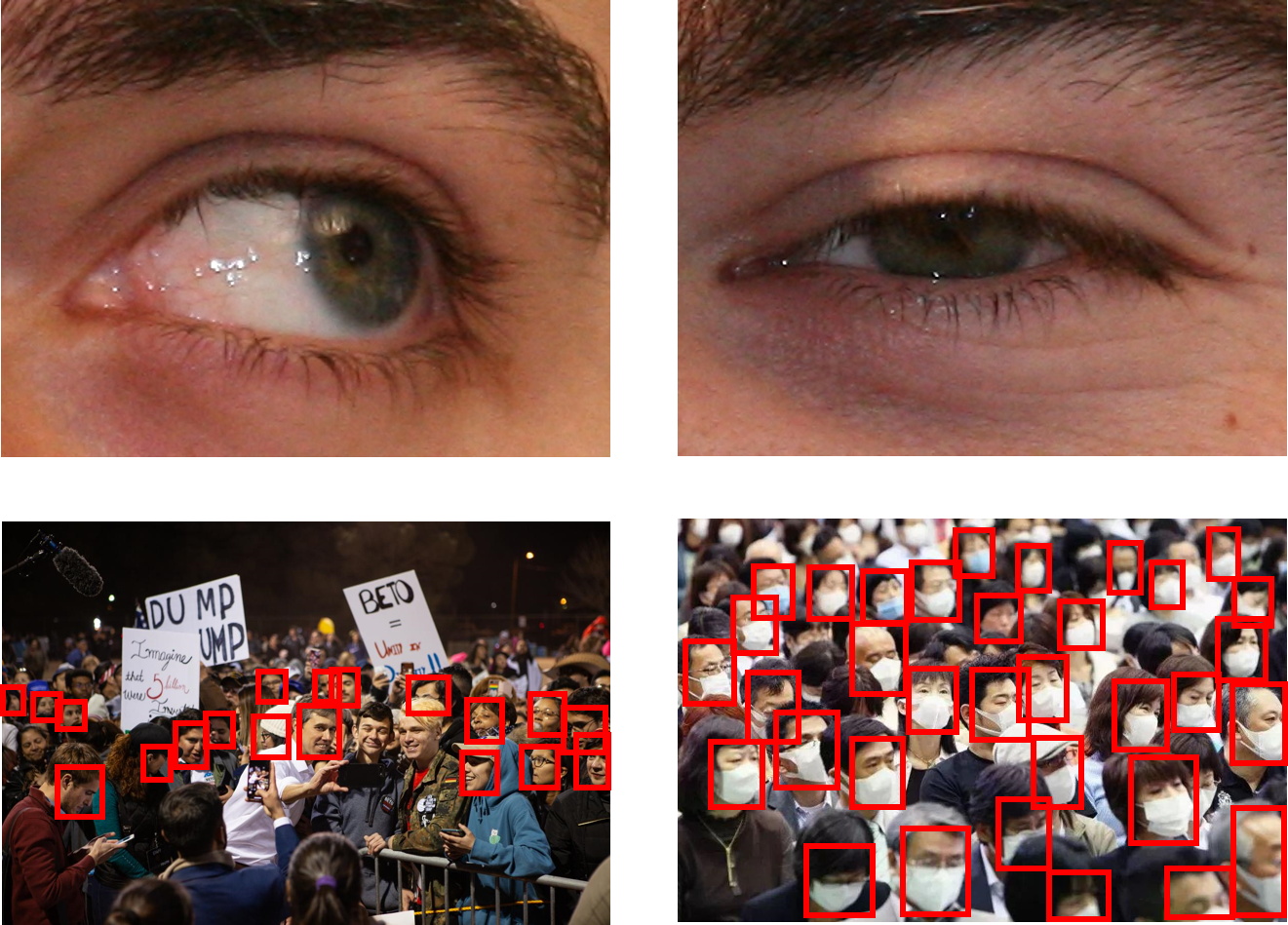}
\end{center}
\setlength{\abovecaptionskip}{0pt}
\setlength{\belowcaptionskip}{0pt}
   \caption{Iris samples and face samples in unconstrained environments.}
\label{fig:occlusion}
\end{figure}

However, tasks of biometrics are different from natural image classification. They are greatly influenced by illumination, expression, occlusion and so on. Among them, occlusion is the most common challenges problem in unconstrained situations~\cite{Ban2013Face} \cite{Azeem2014A} \cite{He2018Dynamic} as shown in Figure~\ref{fig:occlusion}. In these situations, gaps between intra-class samples are significantly enlarged. The generalization abilities of CNNs on occluded cases drops greatly due to the adverse effects of various occlusions.

Masking strategy is commonly seen in literatures for occlusion problems in biometrics. The features of occluded areas are suppressed during matching by masking out the invalid regions. However, detection and labelling of occlusion as masks are error-prone and the errors are accumulated into inaccurate recognition results. More importantly, the features of occluded areas can not be suppressed \emph{after} feature extraction by masks in CNNs which are end-to-end frameworks. And the results of experiments demonstrate that masking the occluded areas \emph{before} feature extraction is harmful.


In this paper, we propose a novel unified framework named Dynamic Graph Representation (DGR) for occluded biometrics. Basically, CNNs are adopted to extract convolutional representations, graph models are leveraged to overcome the occlusion problem. Dynamic graphs are built to remove the nodes of occluded part adaptively. As a result, the similarity between intra-class samples are significantly increased. The discriminative regions and the relationships between the regions are explicitly modeled by graphical blocks in the proposed framework.

Besides, graphical representations provide the reasons behind the decisions. The match or non-match nodes in the proposed dynamic graph method provides a more illustrative and reasonable inference of the  biometrics decision. For example, the probe face image differs from gallery mainly in left eye and nose regions so it is an imposter.

\begin{figure}[t]
\begin{center}
\includegraphics[width=1\linewidth]{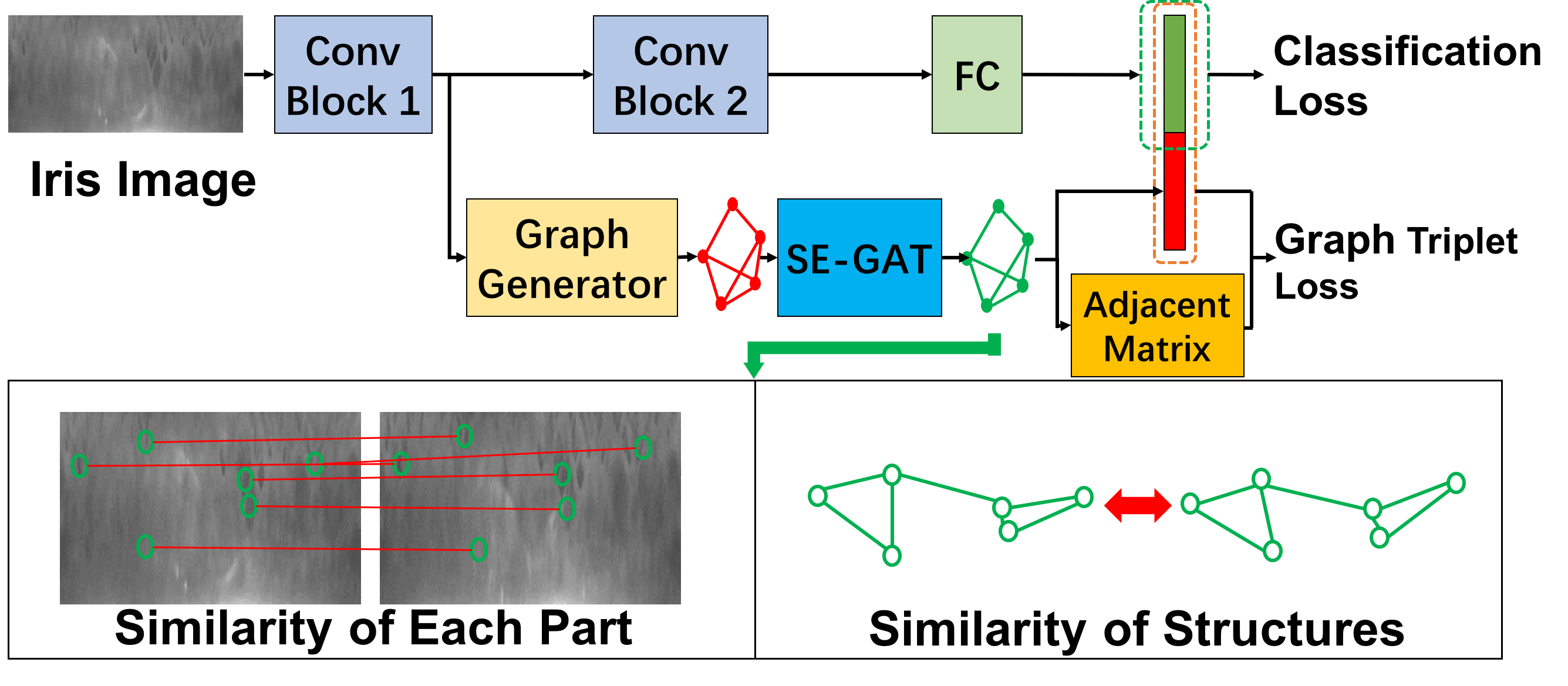}
\end{center}
\setlength{\abovecaptionskip}{0pt}
\setlength{\belowcaptionskip}{0pt}
   \caption{Framework of the proposed method. The similarity of each part of two samples and the similarity of the structures are contained in the Feature Graph.}
\label{fig:framework}
\end{figure}

\begin{figure}[t]
\begin{center}
\includegraphics[width=0.75\linewidth]{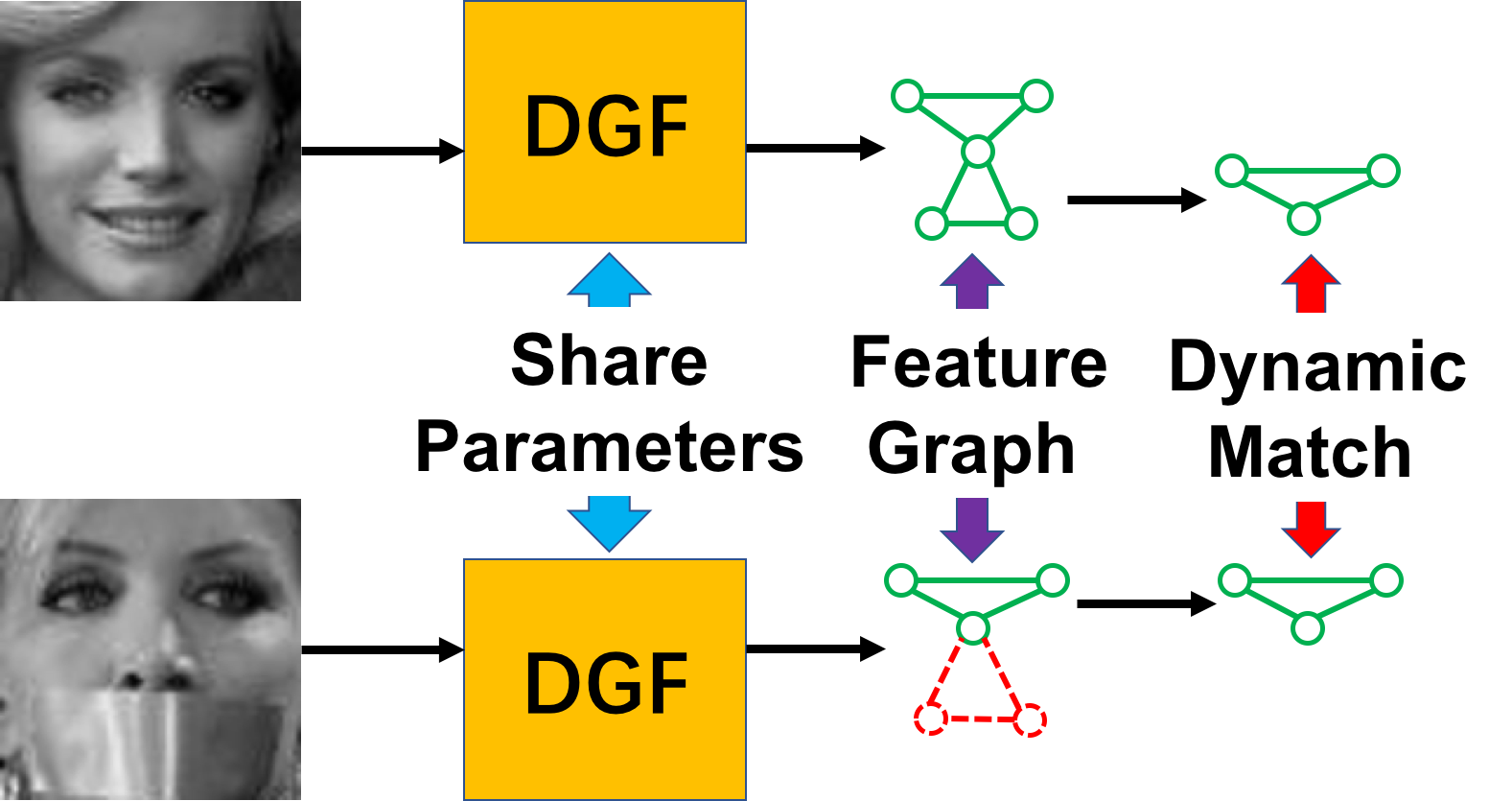}
\end{center}
\setlength{\abovecaptionskip}{0pt}
\setlength{\belowcaptionskip}{0pt}
   \caption{Dynamic matching based on Feature Graphs. The nodes of the occluded parts are removed during matching. Hence, the similarity between intra-class samples are significantly increased.}
\label{fig:match}
\end{figure}

The structure of the proposed framework is shown in Figure~\ref{fig:framework}. Graphical model and CNNs are fused together in the proposed framework: Conv Block 1 produces the feature maps for Graph Generator to construct graphs which named Feature Graphs. Graph Generator could picks up a subset of spatial nodes inside the feature maps, convey edges to express the spatial relationships between nodes and finally combine them together as Feature Graphs as we defined. A novel deep graph model named Squeeze-and-Excitation Graph Attention Networks (SE-GAT) are proposed in this paper to process information of Feature Graphs. During the matching stage, dynamic graphs are built as shown in Figure~\ref{fig:match}. The nodes corresponding to the occluded parts are removed automatically by a straightforward and effective strategy.


In general, Feature Graphs can explicitly show us what parts of two samples are considered and the similarities between them. Hence, the underlying reasons results can be perceived.


The major contributions of this paper can be summarized as follows:

\textbf{1.} To the best of our knowledge, this is the first work that applies deep graphical models for  biometrics. A novel deep graph model called SE-GAT is proposed as an essential component of the proposed framework.

\textbf{2.} The proposed framework provides a novel strategy to deal with occlusions, and the proposed framework achieves better performance by a large margin comparing with baseline methods. The decisions made by our framework are more convincing and reasonable than vanilla CNN-based methods.

\textbf{3.} The proposed framework possesses superior abilities to narrow the gaps between intra-class samples because of the dynamic graph matching, which significantly improves the generalization capability in partially occluded biometrics.





\section{Related Work}
\label{sec:related}

\textbf{Biometric approaches based on CNNs.} Two common traits of biometrics, iris and face recognition, are selected to evaluate the proposed framework. We briefly review researches on these two modalities in this section.

Iris recognition has attracted increasing attention as one of the most accurate and reliable methods for identify authentication. Recently, CNN-based methods for iris recognition are presented. DeepIris is proposed in~\cite{Liu2016DeepIris} for heterogeneous iris matching. Fully convolutional network (FCN) based model named UniNet is applied for iris recognition in~\cite{Zhao2017Towards}. UniNet~\cite{Zhang2018Deep} adopts fully convolutional network (FCN) for feature extraction. MaxoutCNNs is proposed for iris and periocular recognition in~\cite{Zhang2018Deep}. Occlusion problem is common for iris recognition, almost all existing CNN-based methods adopt the masking strategy.

The first method for face representation based on CNN is proposed by Taigman Yaniv et al.~\cite{Taigman2014DeepFace}. A framework employing multiple convolutional networks is proposed by Yi Sun et al.~\cite{Yi2014Deep}. Triplet loss function is applied to get 128-D face embedding representation in~\cite{Schroff2015FaceNet}. A light convolutional architecture is proposed for face recognition in~\cite{Xiang2018A}. Topics about feature space of face draw more attention recently. Numerous of methods for better feature embedding are proposed, including SphereFace~\cite{Liu2017SphereFace}, ArcFace~\cite{Deng2018ArcFace}, CosFace~\cite{Wang2018CosFace} and so on.

There are also region-based or partial-based models for face recognition~\cite{Ou2018Robust} \cite{Cheheb2017Random} \cite{He2018Dynamic}. However, these approaches need to set apart the occluded regions of face images before feature extraction.

\textbf{Graphical methods for biometrics.} There are some graphical methods for biometrics, including iris~\cite{Kerekes2007Graphical} , face~\cite{Kisku2009Probabilistic}, periocular~\cite{Proen2013Periocular}, ear~\cite{Kisku2009Probabilistic} and hand vein~\cite{Horadam2014Hand}. These handcraft methods may provide us useful inspirations, despite these methods can not be blended in deep learning frameworks.

\textbf{Deep learning approaches on graphs.} Graphs are ubiquitous in the real world. Deep learning methods have been extended to graph data recently~\cite{Ziwei2018Deep} \cite{Zhou2018Graph}. Graph Neural Networks (GNNs) are introduced as a recursive neural networks in~\cite{Gori2005A} \cite{Franco2009The} to deal with graph data. Recently, the generalization of convolutional operation draws increasing attention. Researches in this direction can be roughly categorized as spectral approaches and non-spectral approaches~\cite{Zhou2018Graph}. The spectral approaches~\cite{Bruna2014Spectral} \cite{Kipf2016Semi} work with spectral representation of graphs. The non-spectral approaches~\cite{Duvenaud2015Convolutional} \cite{Atwood2016Diffusion} \cite{Hamilton2017Inductive} define convolutional operation directly on graphs. Graph Attention Networks (GAT)~\cite{Veli2017Graph} introduces attention mechanism to graph data. 


\section{Dynamic Graph Representation}
\label{sec:method}

In this section we introduce the framework of Dynamic Graph Representation (DGR). Two crucial components of DGR: Graph Generator and SE-GAT are described separately. After that, a novel loss function designed for the training of Dynamic Graph Feature Learning is shown. Finally, dynamic graph matching strategy is described.

In the proposed framework, images are firstly  sent to a convolutional block (Conv Block 1) as shown in Figure~\ref{fig:framework}. The block consists of several convolutional layers and pooling layers. There are two branches after Conv Block 1. The upper branch in Figure~\ref{fig:framework}  contains another convolutional block (Conv Block 2) and fully connected layers. The two convolutional blocks and the fully connected layers constitute a common CNNs pipeline and any kinds of convolutional neural network can be incorporated into this branch. Typically, global features are extracted at the top of the upper branch.

Another branch starts at the feature maps extracted from the first convolutional block (Conv Block 1). A graph is generated from the feature maps by the Graph Generator. Each node of the graph contains a feature vector and the weights of edges express the relationships between nodes. We call it Feature Graph. Then, Feature Graph is sent to SE-GAT which is a hierarchical feature extractor for graph. SE-GAT is a novel structure based on Graph Attention Networks (GAT) and the details will be shown latter.


Two kinds of loss functions are applied during training. The global feature generated by the fully connected layers is sent to cross-entropy loss function which measures the classification loss. The Feature Graph generated by SE-GAT is sent to a novel graph triplet loss function. The novel loss function provides a new way to measure the similarity of two graphs. Semantic similarity and relationship similarity are both taken into consideration.

\subsection{Graph Generator}

\begin{figure}[t]
\begin{center}
\includegraphics[width=1\linewidth]{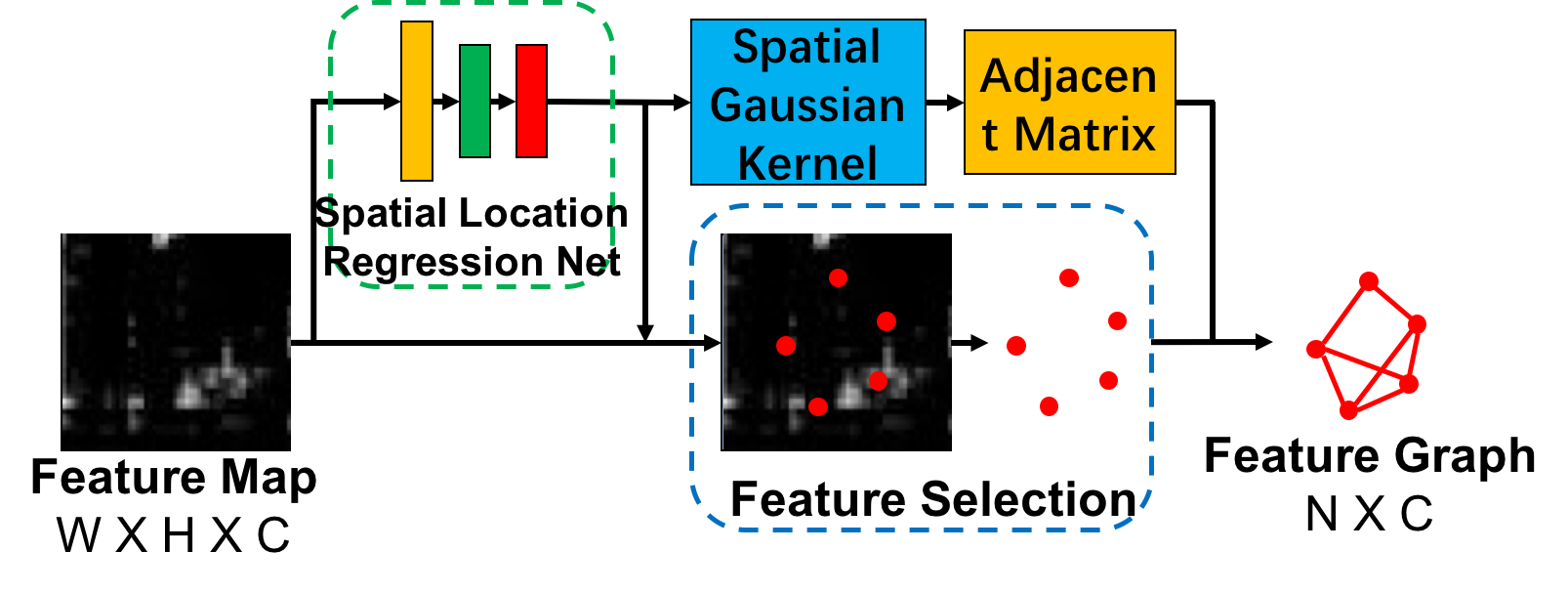}
\end{center}
\setlength{\abovecaptionskip}{0pt}
\setlength{\belowcaptionskip}{0pt}
   \caption{Structure of Graph Generator.}
\label{fig:graphgen}
\end{figure}

There are two steps to generate Feature Graph from feature map as shown in Figure~\ref{fig:graphgen}. The first step is generating the nodes, i.e. regression the spatial location of nodes and sampling from the feature map. The second step is generating the edges, i.e. generating the adjacent matrix according to relationships of nodes.

Spatial location regression of nodes are realized by a light network which called Spatial Location Regression Net as shown in Figure~\ref{fig:graphgen}. This network consists of two convolutional layers and two fully connected layers, which transforms a $W \times H \times C$ feature map into $N$ spatial coordinates in $N \times 2$ format by regression.
 Spatial coordinates are regressed explicitly. Feature vector of each node is sampled from the feature map by bilinear interpolation according to the spatial coordinates:
\begin{equation}
\textbf{f}_{i,j}(c) = \textbf{F}(i,j,c)
\label{equ:sample}
\end{equation}
where $\textbf{f}$ is the feature vector, $\textbf{F}$ is the feature map, $i,~j$ are the spatial coordinates of the node and they are real numbers, $c$ is the index of feature vector and $c \in \{1, 2, ..., C\}$, $C$ is the number of channels of the feature map. Bilinear interpolation is not shown in Equation~\ref{equ:sample} for briefness.

The adjacent matrix is generated by Gaussian kernel function according to the spatial position of nodes: 
\begin{equation}
\label{equ:gaussian}
\textbf{M}_{adj}(a,~b)=\left\{
\begin{array}{cl}
0 & ||\textbf{n}_a-\textbf{n}_b||_2>R, \\
exp(-\frac{||\textbf{n}_a-\textbf{n}_b||_2^2}{2 R^2}) & ||\textbf{n}_a-\textbf{n}_b||_2<R.
\end{array}
\right.
\end{equation}
where $\textbf{M}_{adj}$ is the $N\times N$ adjacent matrix, $N$ is the number of nodes, $\textbf{n}_a,~\textbf{n}_b$ are the spatial coordinates of two nodes, $R$ is the scale of receptive field of each node. According to Equation~\ref{equ:gaussian}, if the receptive fields on input image of two nodes are overlapped, edge will be generated between them. And the larger the overlapped area is, the larger the weight of the edge is.

After the two steps, $N \times C$ Feature Graph is generated from $W \times H \times C$ feature map, where $N$ is the number of nodes. The Spatial Location Regression Net selects the most important locations of feature map as nodes of the Feature Graph. Spatial Gaussian kernel function establishes relationships between nodes with overlapped receptive fields.


\subsection{SE-GAT}

SE-GAT is a new network based on Graph Attention Networks (GAT)~\cite{Veli2017Graph}. It consists of two kinds of layers: squeeze-and-excitation (SE) layer and graph attention (GAT) layer.

\begin{figure}[t]
\begin{center}
\includegraphics[width=1\linewidth]{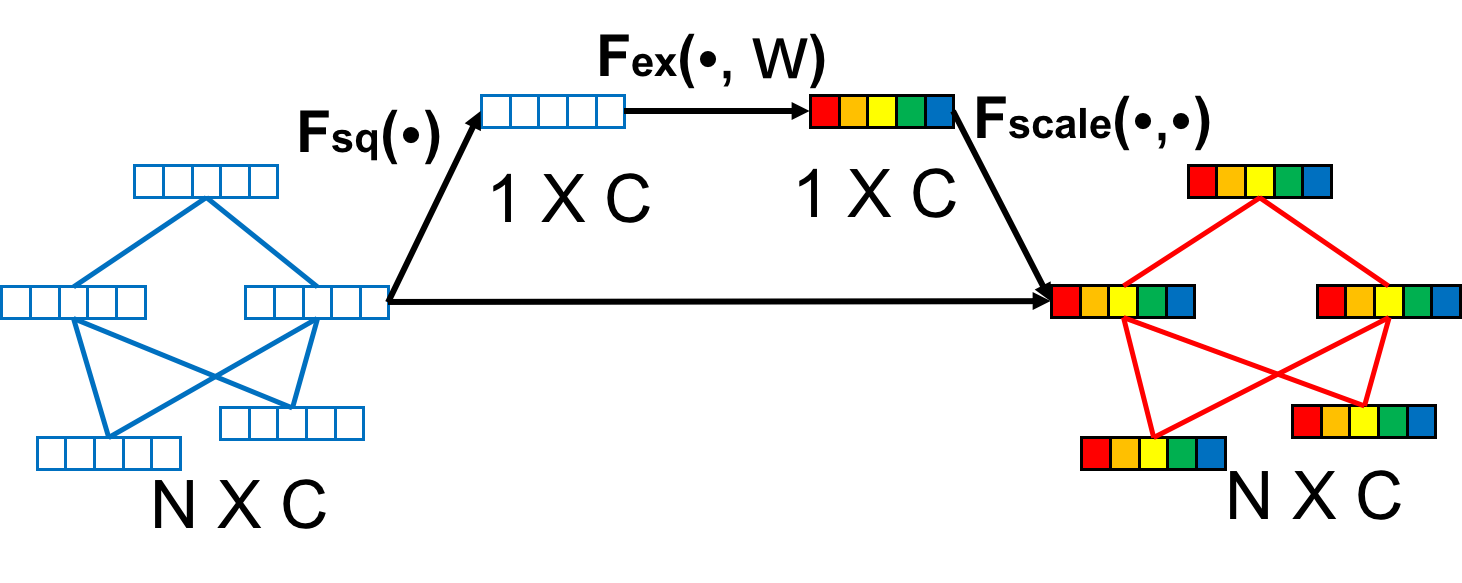}
\end{center}
\setlength{\abovecaptionskip}{0pt}
\setlength{\belowcaptionskip}{0pt}
   \caption{Structure of SE layer.}
\label{fig:selayer}
\end{figure}

Squeeze-and-excitation layer is a generalization of squeeze-and-excitation block~\cite{Hu2018Squeeze}. As shown in Figure~\ref{fig:selayer}, the average energy of each channel of nodes is calculated by the global pooling operation:
\begin{equation}
\textbf{z}(c) = F_{sq}(\textbf{G})= \frac{1}{N}\sum_{i=1}^N \textbf{f}_i(c)
\label{equ:squeeze}
\end{equation}
where $\textbf{z}$ is the squeezed vector, $\textbf{G}$ is the Feature Graph, $N$ is the number of nodes, $c$ is the index of the feature vector of each node and $c \in \{1, 2, ..., C\}$. Two fully connected layers are used to make use of the global information:
\begin{equation}
\textbf{s} = F_{ex}(\textbf{z},~\textbf{W}) = \delta_2(\textbf{W}_2\delta_1(\textbf{W}_1 z)) 
\label{equ:excitation}
\end{equation}
where $\textbf{s}$ is the scale vector, $\textbf{W}_1,~\textbf{W}_2$ are parameters of the two fully connected layers, $\delta_1,~\delta_2$ are ReLU function. The scale vector $\textbf{s}$ is used to re-scale each channel of the Feature Graph:
\begin{equation}
\textbf{f}'_i(c) = \textbf{s}(c)\textbf{f}_i(c)
\label{equ:rescale}
\end{equation}
where $\textbf{f}'$ is the re-scaled feature vector, $c$ is the index of feature vector and $c \in \{1, 2, ..., C\}$, $i$ is the index of nodes and $i \in \{1, 2, ..., N\}$.

The graph attention (GAT) layer is a modified variant of the layer in~\cite{Veli2017Graph}. The weights of different edges of the Feature Graph is different rather than the same. The formulation of the layer is given as follows:
\begin{equation}
\textbf{f}_a' = \delta (\alpha_{ab}(a,b) \textbf{W} \textbf{f}_b)
\label{equ:modgat}
\end{equation}

\begin{equation}
\alpha_{ab} = \frac{exp(\delta(\textbf{M}_{adj}(a,b)e_{ab})}{ \sum_k exp(\delta(\textbf{M}_{adj}(a,k)e_{ab}))}
\label{equ:alpha}
\end{equation}

\begin{equation}
e_{ab} = \textbf{w}_{att}^T[\textbf{Wf}_a || \textbf{Wf}_b]
\label{equ:e}
\end{equation}
where $\textbf{f}'\in \mathbb{R}^{C}$ is the output vector, $\textbf{f}\in \mathbb{R}^{C}$ is the input vector, $\delta$ is nonlinear activation function, $\alpha_{ab}$ is the attention value between node $a$ and node $b$, $\textbf{W}\in \mathbb{R}^{C\times C}$ and $\textbf{w}_{att}\in \mathbb{R}^{2C}$ are the parameters of the layer, $\textbf{M}_{adj}$ is the adjacent matrix of input graph, $||$ is the concatenation operation. 

\begin{figure}[t]
\begin{center}
\includegraphics[width=0.9\linewidth]{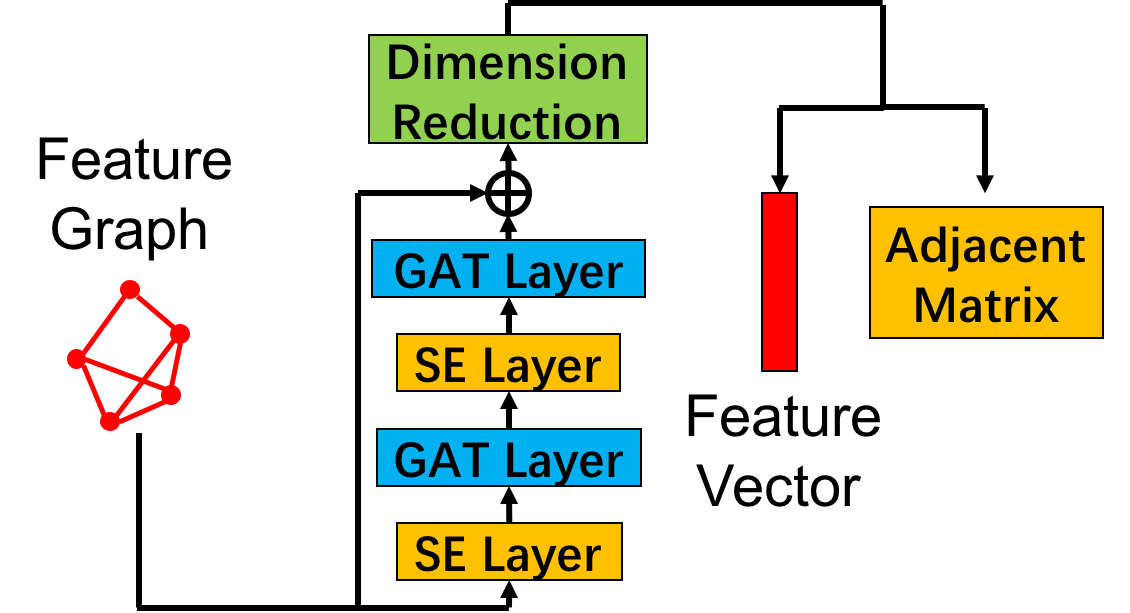}
\end{center}
\setlength{\abovecaptionskip}{0pt}
\setlength{\belowcaptionskip}{0pt}
   \caption{Framework of SE-GAT.}
\label{fig:segat}
\end{figure}

The framework of SE-GAT is shown in Figure~\ref{fig:segat}. Residual structure has been proved to be effective for CNNs~\cite{He2016Deep}. We extend it to graph model in SE-GAT. There are four layers, two of them are SE layer and the rest are GAT layer as shown in Figure~\ref{fig:segat}. Then the output is added to the input before a dimension reduction layer. The dimension reduction layer reduces the dimension of feature vectors of nodes:
\begin{equation}
\textbf{f}_i' = \delta(\textbf{W}_{C'\times C}\textbf{f}_i)
\label{equ:reduce}
\end{equation}
where $\textbf{W}_{C'\times C}$ is the parameter matrix, $\delta$ is nonlinear activation function, $C'$ is the dimension of output feature vectors. Two kinds of features are separated from the output Feature Graphs of SE-GAT: feature vectors which express the semantic feature, adjacent matrix which expresses the relationship feature.

\subsection{Graph Triplet Loss Function}
\label{sec:loss}
We describe a new approach to measure the similarity between two Feature Graphs firstly. Then, the formula of the novel loss function is given.

The output of DGR contains two kinds of information: semantic information in the feature vectors and spatial relationship information in the adjacent matrixes of Feature Graphs. Hence, the similarity measure of two Feature Graphs consists of two terms:
\begin{equation}
\mathit{S} = \mathit{S}_{fea} + \mathit{S}_{adj}
\label{equ:sim}
\end{equation}
where $\mathit{S}_{fea}$ is the semantic similarity, $\mathit{S}_{adj}$ is the relationship similarity. 
\begin{equation}
\mathit{S}_{fea} = cosine(\widetilde{\textbf{f}_1}, \widetilde{\textbf{f}_2})
\label{equ:feasim}
\end{equation}
where $cosine$ represents the cosine similarity, $\widetilde{\textbf{f}_1}, \widetilde{\textbf{f}_2}$ are the concatenation of all feature vectors of Feature Graphs and feature vectors generated by fully connected layers.
\begin{equation}
\mathit{S}_{adj} = ||\textbf{M}_{adj}^1 - \textbf{M}_{adj}^2||_2
\label{equ:adjsim}
\end{equation}
where $\textbf{M}_{adj}^1, \textbf{M}_{adj}^2$ are the adjacent matrix of Feature Graphs.

The graph triplet loss is:
\begin{equation}
\mathit{L} = max\{0,m + S_{anc-neg} - S_{anc-pos}\}
\label{equ:triploss}
\end{equation}
where $S_{anc-neg}$ is the similarity between anchor sample and negative sample, $S_{anc-pos}$ is the similarity between anchor sample and positive sample, $m$ is a positive margin. We set $m=1$ during training.


\subsection{Dynamic Graph Match}
For two Feature Graphs, $G_A$ and $G_B$, the mean of cosine similarities of corresponding nodes are calculated:
\begin{equation}
s_{gate} = \frac{1}{N}\sum_{i=1}^N cosine(\textbf{f}_i^A,~\textbf{f}_i^B)
\label{equ:gate}
\end{equation}
where $\textbf{f}_i^A$ and $\textbf{f}_i^B$ are the feature vectors of corresponding nodes, $N$ is the number of nodes in each Feature Graph. If the similarity of a node pair is less than $s_{gate}$, these two nodes are removed from the two Feature Graphs:
\begin{equation}
G_A^{Dyn}, G_B^{Dyn} = Re(G_A, G_B, s_{gate})
\label{equ:remove}
\end{equation}
where $Re$ represents the remove operation according to the $s_{gate}$. The dynamic graphs $G_A^{Dyn}, G_B^{Dyn}$ are built and the similarity of $G_A^{Dyn}, G_B^{Dyn}$ is calculated according to the approach described above.

The node pairs with lower similarities are removed because the variances of occluded parts are much larger than the visible parts. And the gate given by the mean of similarities provides a straightforward and effective way to built dynamic graphs.


\section{Experiments}
\label{sec:experiment}

Two common modalities of biometrics, iris and face, are selected to evaluate the proposed method.



\subsection{Experiments of Iris Recognition}

\subsubsection{Protocols and Databases}
A simple novel convolutional architecture is used for iris experiments. The Conv Block 1 contains four convolutional layers and the Conv Block 2 contains one convolutional layer.
 The number of nodes of Feature Graph is 32.

For all iris images, the preprocessing procedure contains three steps: 1) eye detection by Haar-like Adaboost detectors~\cite{Viola2004Robust}, 2) iris boundaries localization using method in~\cite{Zhaofeng2009Toward}, 3) iris normalization by rubber sheet model~\cite{Daugman1993High}. Iris images are normalized to a rectangle with $128 \times 256$ resolution.

Four databases are used for experiments: (1) ND CrossSensor Iris 2013 Dataset-LG4000. 
It contains 29,986 iris samples from 1,352 classes. (2) CASIA Iris Image Database V4-Distance. Images of this database are acquired from 3 meters away. It contains 2,446 iris samples from 284 classes. (3) CASIA-Iris-M1-S2. Images of this database are acquired by mobile devices. It contains 6,000 iris samples from 400 classes. (4) CASIA Iris Image Database V4-Lamp. This database contains 16,212 iris samples from 819 classes. A lamp was turned on/off close to the subject to introduce elastic deformation under different illumination conditions. 

The images of left eyes are selected for training and the images of right eyes for testing.


\subsubsection{Comparisons to State-of-the-Art}
The proposed method is compared with the state-of-the-art approaches, including log-Gabor~\cite{L2003}, Ordinal Measures (OMs)~\cite{Zhenan2009Ordinal}, UniNet~\cite{Zhao2017Towards} and MaxoutCNNs~\cite{Zhang2018Deep}.

\begin{figure}[t]
\vspace{-0.2cm}
\begin{center}
\includegraphics[width=1\linewidth]{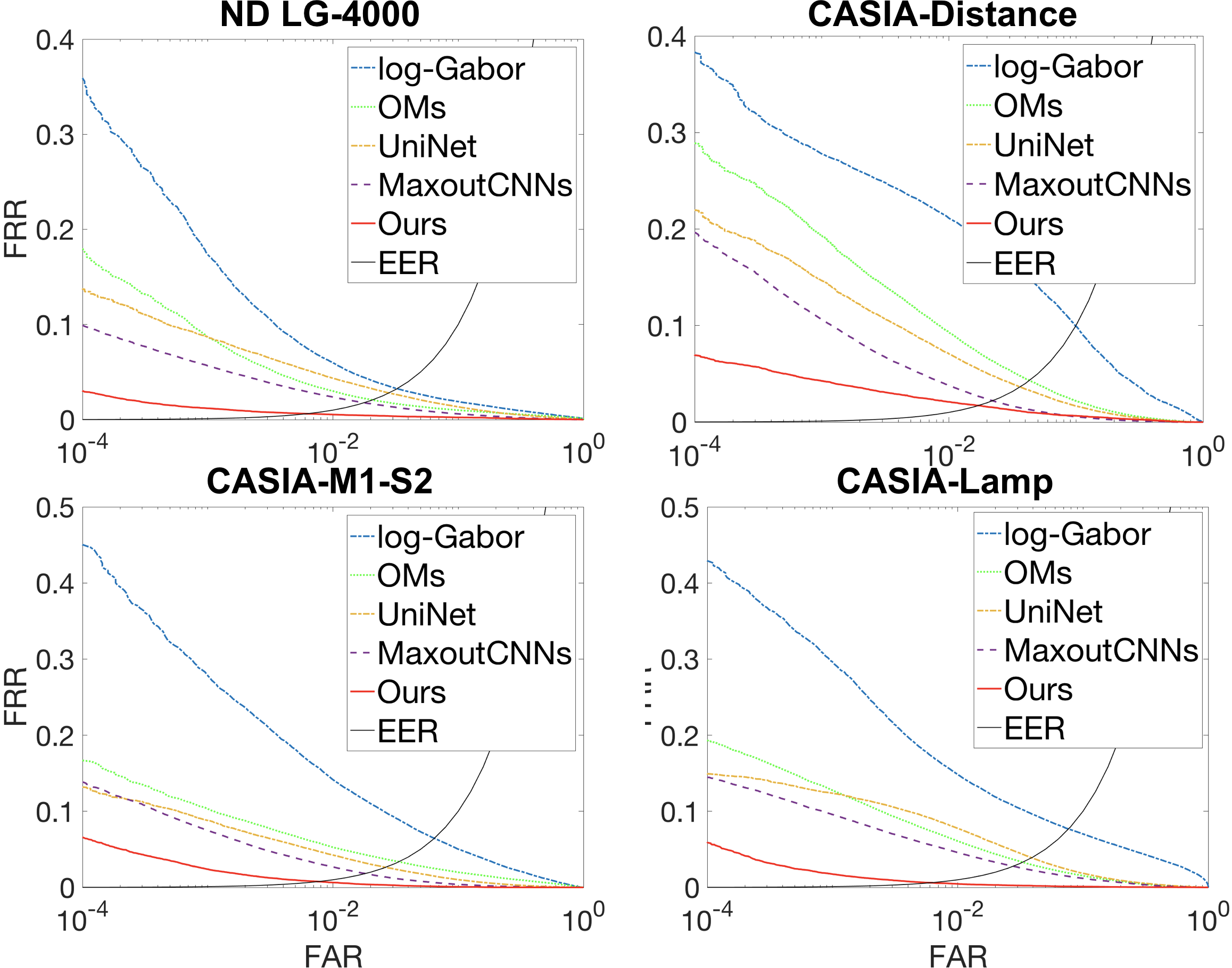}
\end{center}
\setlength{\abovecaptionskip}{0pt}
\setlength{\belowcaptionskip}{0pt}
   \caption{DET curves of iris recognition experiments.}
\label{fig:roc_iris}
\vspace{-0.1cm}
\end{figure}

The results are shown in Figure~\ref{fig:roc_iris} and Table~\ref{tab:iris}. Significant improvements from proposed method can be found on all four databases.

\begin{table*}
\begin{center}
\setlength{\abovecaptionskip}{0pt}
\setlength{\belowcaptionskip}{0pt}
\caption{False reject rates (FRR) and equal error rates (EER) of iris recognition experiments.}
\label{tab:iris}
\setlength{\tabcolsep}{0.1mm}
{
\begin{tabular}{c|c|c|c|c|c|c|c|c}
\hline
\multirow{2}*{~} & \multicolumn{2}{|c|}{\bf{ND-LG4000}} & \multicolumn{2}{|c|}{\bf{CASIA-Distance}} & \multicolumn{2}{|c|}{\bf{CASIA-M1-S2}} & \multicolumn{2}{|c}{\bf{CASIA-Lamp}} \\
 \cline{2-9}
& FRR@FAR=0.01\% & EER & FRR@FAR=0.01\% & EER & FRR@FAR=0.01\% & EER & FRR@FAR=0.01\% & EER \\
\hline
log-Gabor & 35.89\% & 3.25\% & 38.30\% & 9.96\% & 45.02\% & 6.35\% & 42.94\% & 7.64\% \\
\hline
OMs & 17.94\% & 2.07\% & 28.87\% & 4.31\% & 16.67\% & 3.35\% & 19.32\% & 3.30\% \\
\hline
UniNet & 13.75\% & 2.80\% & 21.93\% & 3.63\% & 13.18\% & 2.64\% & 14.93\% & 3.88\%  \\
\hline
MaxouCNNs & 9.90\% & 1.77\% & 19.61\% & 2.21\% & 13.83\% & 1.81\% & 14.46\% & 2.77\%  \\
\hline
Ours & \bf{3.02\%} & \bf{0.62\%} & \bf{6.94\%} & \bf{1.71\%} & \bf{6.57\%} & \bf{0.76\%} & \bf{5.92\%} & \bf{0.61\%} \\
\hline
\end{tabular}}
\end{center}
\end{table*}

\begin{figure}[h]
\begin{center}
\includegraphics[width=0.5\linewidth]{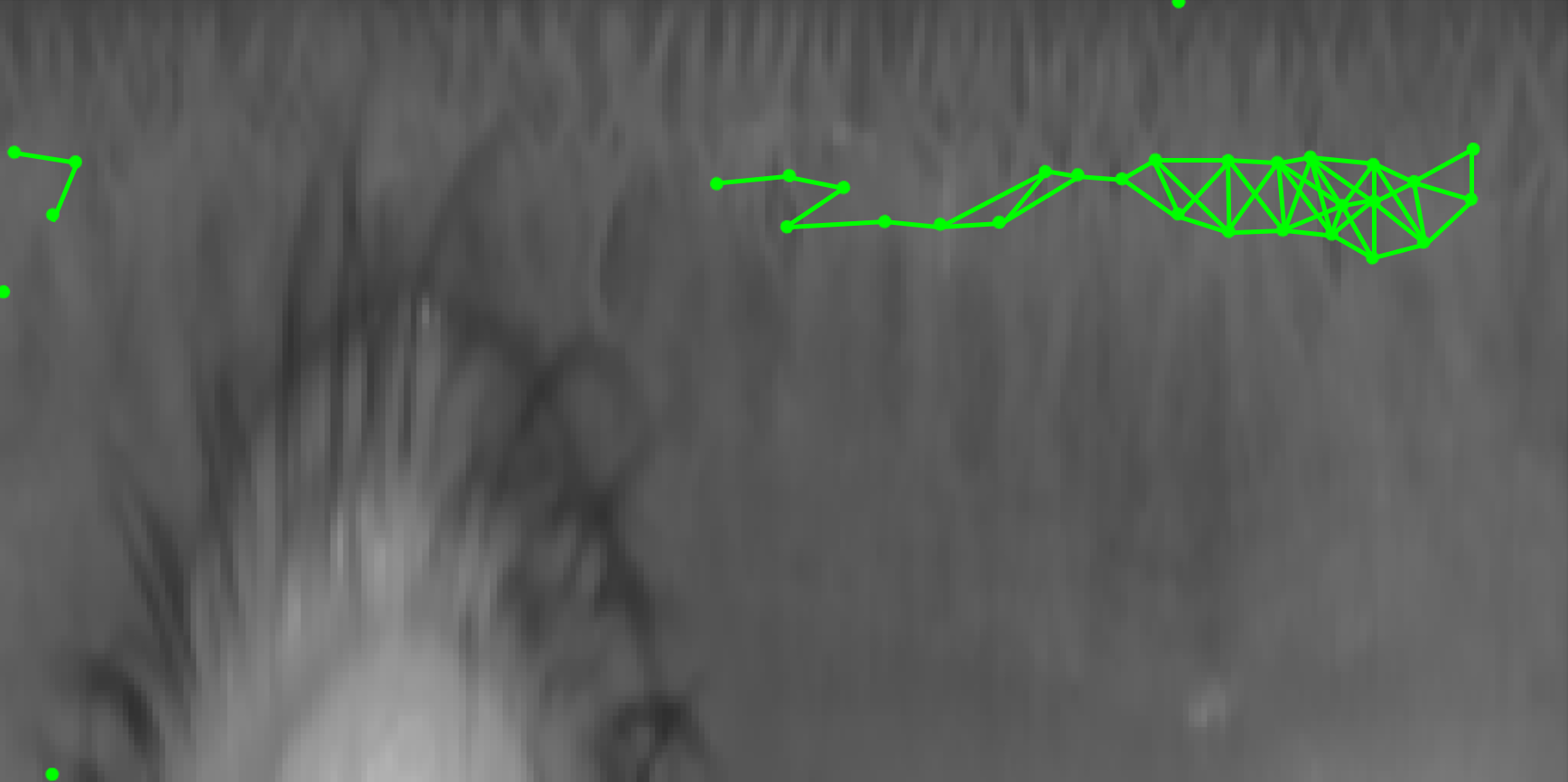}
\end{center}
\setlength{\abovecaptionskip}{0pt}
\setlength{\belowcaptionskip}{0pt}
   \caption{Visualization of the Feature Graph of a iris sample from the CASIA Iris Image Database V4-Lamp.}
\label{fig:iris_graph}
\end{figure}

Feature Graph of a iris sample from the CASIA Iris Image Database V4-Lamp is visualized in Figure~\ref{fig:iris_graph}. The nodes avoid the eyelid which is useless for iris recognition as we can see. Almost all nodes locate around the boundary of pupillary zone and ciliary zone. The locations of nodes indicate that the network ``thinks'' this area is the most important part of iris for recognition.

\subsubsection{Occluded Iris Recognition}
\label{exp:occ_iris}
\begin{figure}[h]
\begin{center}
\includegraphics[width=0.8\linewidth]{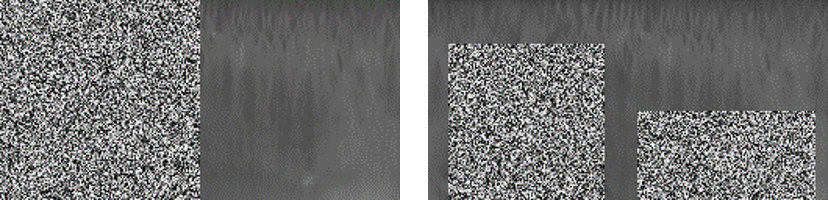}
\end{center}
\setlength{\abovecaptionskip}{0pt}
\setlength{\belowcaptionskip}{0pt}
   \caption{Two situations of occluded iris samples.}
\label{fig:iris_occ}
\end{figure}

Occluded iris recognition experiment is launched on the test database of the ND CrossSensor Iris 2013 Dataset-LG4000. Special area of iris samples are covered by random noise to simulate the occluded situations in Figure~\ref{fig:iris_occ}. Two kinds of occlusion situations shown in Figure~\ref{fig:iris_occ} are selected randomly for a special iris sample pair while the percentage of covered area remains. Note that all models used in this experiment are not trained on the occluded database.

\begin{figure}[h]
\begin{center}
\includegraphics[width=0.7\linewidth]{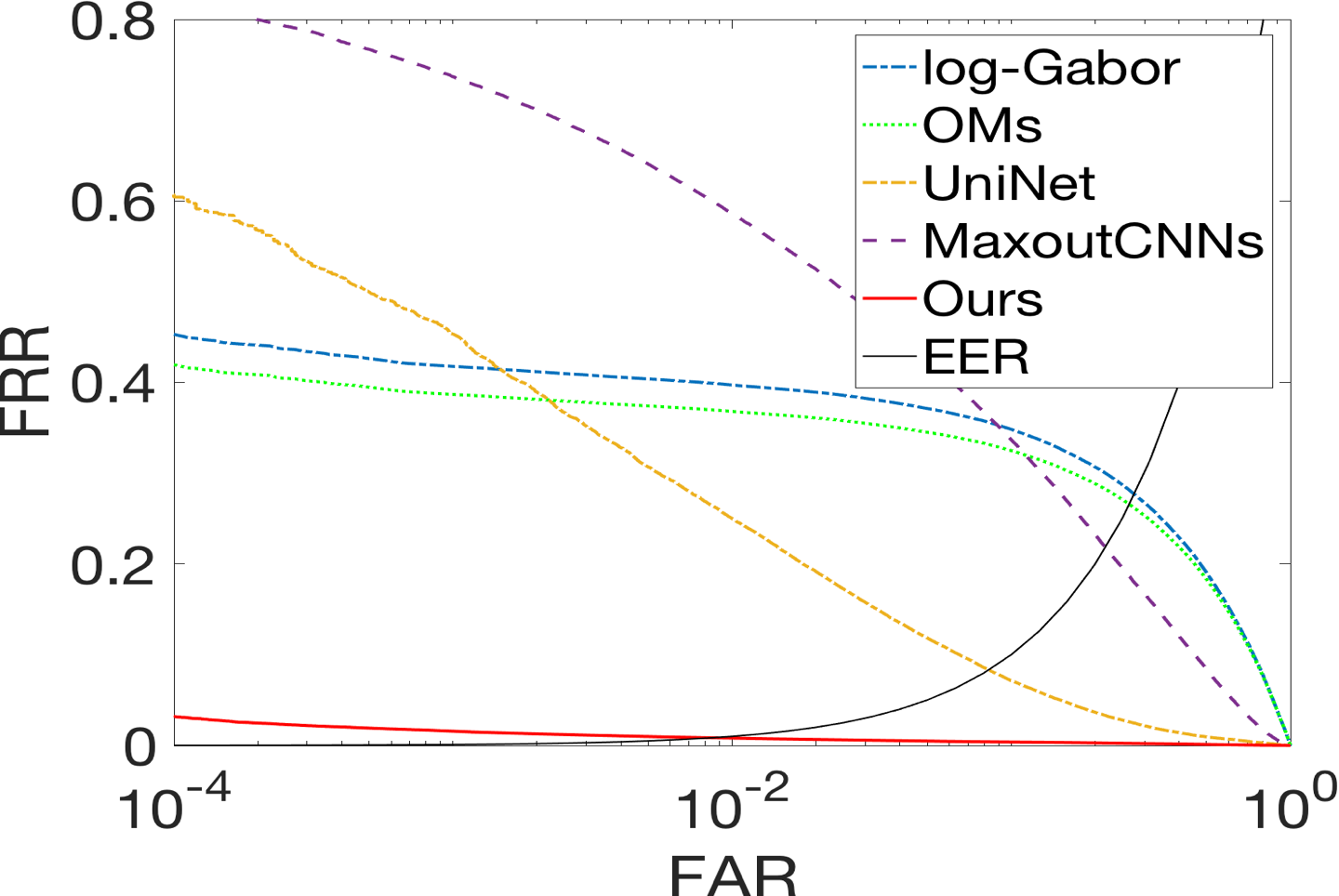}
\end{center}
\setlength{\abovecaptionskip}{0pt}
\setlength{\belowcaptionskip}{0pt}
   \caption{DET curves under occluded situations. The percentage of occluded area is $30\%$.}
\label{fig:roc_occ}
\end{figure}

The results of $30\%$ occluded are shown in Figure~\ref{fig:roc_occ} and Table~\ref{tab:roc_occ}. The performance of the proposed method under occluded situations reduces slightly while the other method become worse seriously. The results indicate that the propose method has desired generalization ability for occluded situations.

\begin{table}
\begin{center}
\setlength{\abovecaptionskip}{0pt}
\setlength{\belowcaptionskip}{0pt}
\caption{FRR and EER of occluded situations. The percentage of occluded area is $30\%$.}
\label{tab:roc_occ}
\setlength{\tabcolsep}{3mm}
{
\begin{tabular}{c|c|c}
\hline
& FRR@FAR=0.01\% &EER \\
\hline
log-Gabor & 45.29\% & 27.26\%  \\
\hline
OMs & 41.94\% &26.29\%  \\
\hline
UniNet & 60.66\% & 8.31\%  \\
\hline
MaxoutCNNs & 81.75\% & 21.79\%  \\
\hline
Ours & \bf{3.22\%} & \bf{0.85\%}  \\
\hline
\end{tabular}}
\end{center}
\end{table}

\begin{figure}[h]
\begin{center}
\includegraphics[width=0.68\linewidth]{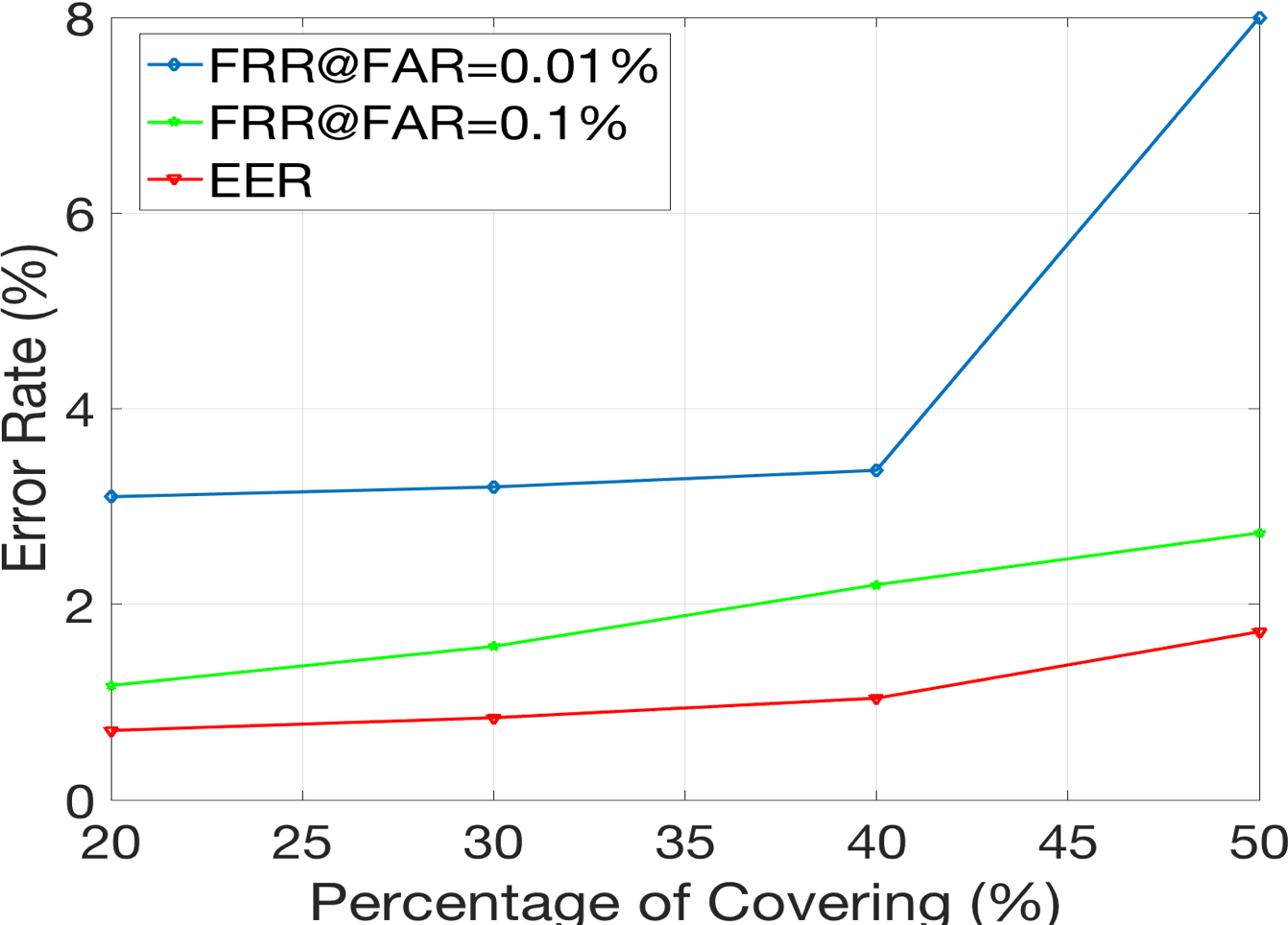}
\end{center}
\setlength{\abovecaptionskip}{0pt}
\setlength{\belowcaptionskip}{0pt}
   \caption{Performance of the proposed method under different percentages of occlusion.}
\label{fig:occ}
\end{figure}

Additional experiment are conducted to evaluate the influence of the covered area size. Performance of the proposed method under different percentages of covering are provided in Figure~\ref{fig:occ}.


\subsection{Experiments of Occluded Face Recognition}

\subsubsection{Protocols and Database}
Light CNN ~\cite{Xiang2018A} is adopted to build the architecture for face recognition. The Light CNN-9 version which contains 9 convolutional layers is selected. The Conv Block 1 consists of the first 5 convolutional layers, and the Conv Block 2 consists of rest 4 convolutional layers. The number of nodes of Feature Graph is 64. Light CNN-9 is selected as the baseline method for comparison naturally.

The CASIA-WebFace~\cite{Yi2014Learning} is adopted as the training database. The CASIA-WebFace is a public database which contains 494,414 face images from 10,575 subjects. All face images in training database are converted to gray-scale images before normalized to $128\times 128$ resolution according to landmarks. 
Horizontal mirror operation is conducted for data augmentation because faces have nearly symmetric structure.


\subsubsection{Occluded Face Verification}
\label{exp:lfw}
\begin{figure}[h]
\begin{center}
\includegraphics[width=0.7\linewidth]{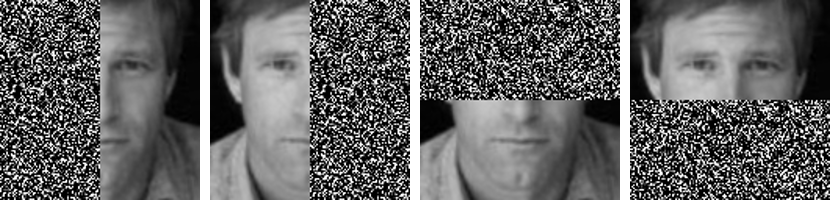}
\end{center}
\setlength{\abovecaptionskip}{0pt}
\setlength{\belowcaptionskip}{0pt}
   \caption{Four situations of occluded face samples.}
\label{fig:face_occ}
\end{figure}

A simulated occluded face database named Occluded-LFW which is based on the Labeled Faces in the Wild (LFW) database is used for evaluation. LFW contains 13,233 images from 7,749 individuals. After the same pre-processing of the training images, special area of face images are covered by random noise to simulate the occluded situations during this experiment as shown in Figure~\ref{fig:face_occ}. Four kinds of occlusion situations shown in Figure~\ref{fig:face_occ} are selected randomly for a special face image pair while the percentage of covered area remains. In order to evaluate the generalization ability of the proposed method, the models in this experiment are not tuned on the occluded database.

We follow the Labeled Faces in the Wild (LFW) benchmark protocol\footnote{http://vis-www.cs.umass.edu/lfw/pairs.txt}, where 3,000 positive pairs and 3,000 negative pairs of images are selected for face verification. One of the face image of each pair is from the Occluded-LFW database and the other is from the LFW database.

\begin{figure}[h]
\begin{center}
\includegraphics[width=1.0\linewidth]{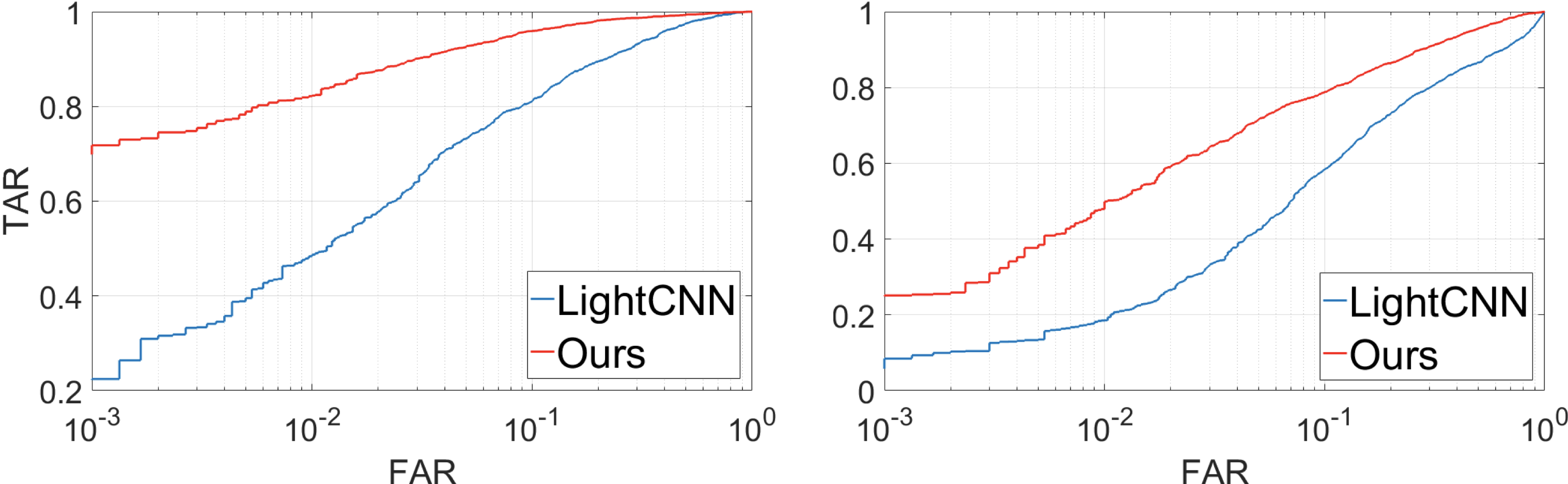}
\end{center}
\setlength{\abovecaptionskip}{0pt}
\setlength{\belowcaptionskip}{0pt}
   \caption{ROC curves of occluded face verification. The percentage of occluded area of the left figure is 30\%. The percentage of occluded area of the right figure is 50\%.}
\label{fig:face_ver}
\end{figure}

Experiments with two different percentages of occluded area are launched: 30\% and 50\%. The results are shown in Figure~\ref{fig:face_ver} and Table~\ref{tab:lfw}. The performance of the  proposed method is better than the baseline method obviously on both two percentages of occluded area.

\begin{table}
\begin{center}
\setlength{\abovecaptionskip}{0pt}
\setlength{\belowcaptionskip}{0pt}
\caption{Results of occluded face verification on the Occluded-LFW.}
\label{tab:lfw}
\setlength{\tabcolsep}{1mm}
{
\begin{tabular}{c|c|c}
\hline
& TPR@FTR=0.1\% &  100\%-EER \\
\hline

 \multicolumn{3}{c}{Percentage of occlusion: 30\%}\\
\hline
LCNN-9 &  22.47\% & 85.97\%  \\
\hline
Ours & \textbf{71.8\%} & \textbf{93.63\%}  \\
\hline

 \multicolumn{3}{c}{Percentage of occlusion: 50\%}\\
\hline
LCNN-9  & 8.47\% & 76.47\%  \\
\hline
Ours &  \textbf{25.19\%} & \textbf{84.23\%}  \\
\hline

\end{tabular}}
\end{center}
\end{table}

\begin{figure}[h]
\begin{center}
\includegraphics[width=0.55\linewidth]{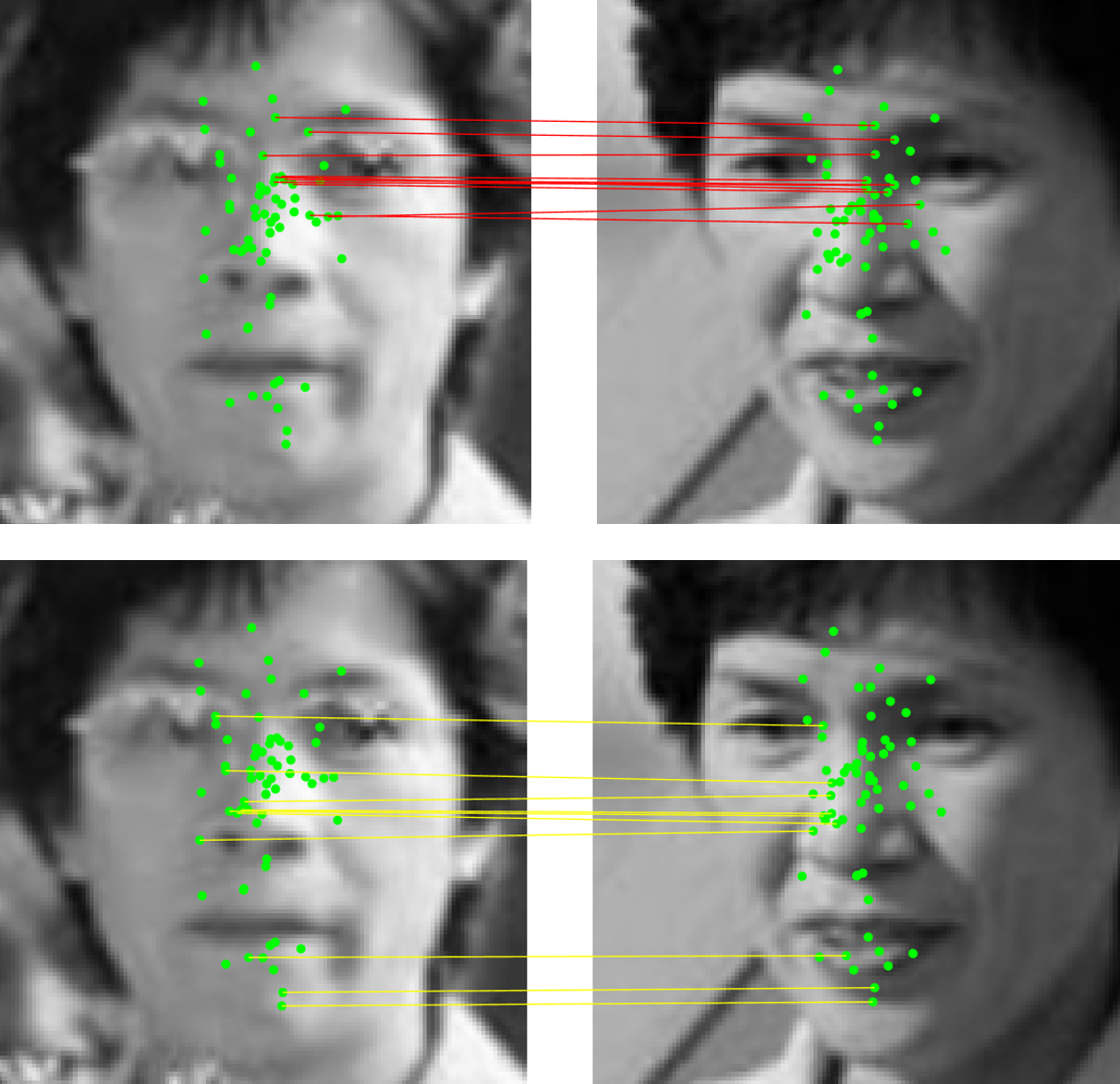}
\end{center}
\setlength{\abovecaptionskip}{0pt}
\setlength{\belowcaptionskip}{0pt}
   \caption{Matched nodes of a pair of images. The first row shows the 10 node pairs with highest similarity scores. The second row shows the 10 node pairs with lowest similarity scores.}
\label{fig:face_match}
\end{figure}

Feature Graphs of a pair of face images are visualized in Figure~\ref{fig:face_match}. The first row shows the 10 pairs of nodes with highest similarity scores. The second row shows the 10 node pairs with lowest similarity scores. All the 10 pairs nodes with highest similarity scores lie in the left area of the right eye which indicates that the network ``thinks'' this is the most similar part of the two images. 
This example proves that we can read out the reasons for the results shown by the network from the Feature Graph. 

\subsection{Experiments on Masking Strategy}
\label{exp:mask}

In this section, experiments on masking strategy are launched for comparison.
Experiments of this section are launched on the test database of the ND CrossSensor Iris 2013 Dataset-LG4000. 
The percentage of occluded are of each sample is 50\%. Two frameworks are used to evaluation the masking strategy: the proposed method and the backbone, which is a normal CNNs, of it. In the situation without masking strategy, the occluded samples are sent to networks directly. In the situation with masking strategy, occluded areas of samples are masked by precise masks, which eliminate the influence of inaccuracy of masks, before sent to networks.

\begin{figure}[h]
\begin{center}
\includegraphics[width=0.7\linewidth]{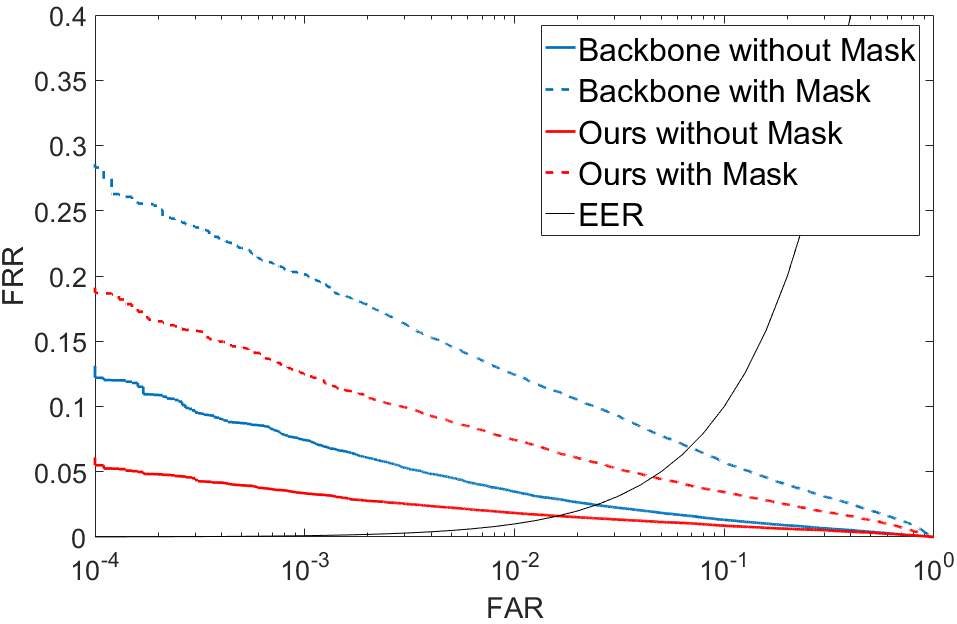}
\end{center}
\setlength{\abovecaptionskip}{0pt}
\setlength{\belowcaptionskip}{0pt}
   \caption{Comparisons to masking strategy. The percentage of occluded area is 50\%.}
\label{fig:mask}
\end{figure}

The results are shown in Figure~\ref{fig:mask}. The performances of situations with masking strategy are worse than the performances without masking strategy in both two frameworks as we can observe in Figure~\ref{fig:mask}. The reasons may be that the masks disturb the feature extraction. 
The results indicate that covering occluded areas of input images by masks is harmful to deep learning framework.


\subsection{Parameter Analysis}
\label{exp:parameter}

The number of nodes of Feature Graph is a important hyper-parameter of the proposed framework. To evaluate the influence of the number of nodes, we conduct experiments on different number of nodes. The test database of the ND CrossSensor Iris 2013 Dataset-LG4000 are selected to launch the experiments.

The results are shown in Table~\ref{tab:node}. The dimension of feature vectors are also listed. Too many nodes will increase computational cost and some distractive information is included. On the other hand, too few nodes fail to achieve good accuracy. An analysis on the graph nodes/connections will be added.

\begin{table}[h]
\begin{center}
\setlength{\abovecaptionskip}{0pt}
\setlength{\belowcaptionskip}{0pt}
\caption{Results on the ND-LG4000}
\label{tab:node}
\setlength{\tabcolsep}{0.1mm}
{
\begin{tabular}{c|c|c|c}
\hline
Num Nodes& FRR@FAR=0.01\% &EER&Feature Dimension \\
\hline
8 & 4.51\% & 0.93\%&512  \\
\hline
32 & \bf{3.01}\% & \bf{0.62}\%&1280  \\
\hline
128 & 4.56\% & 0.73\% &4352 \\
\hline
\end{tabular}}
\end{center}
\vspace{-0.8cm}
\end{table}

\section{Conclusion}
\label{sec:conclude}

We propose a novel deep learning framework called Dynamic Graph Representation (DGR) for occlusion handling in biometrics in this paper. Dynamic graphs are adopted to overcome the occlusion situations. 
A novel deep graph model is proposed for processing of the Feature Graph.
The excellent performance and superior generalization ability are demonstrated by extensive experiments on multiple modality databases. 
The main idea of this paper is straightforward, and we believe that there is much room for improvements. Further exploration of this work should focus on the better graph generation strategy and the better method for dynamic graph building.

\section*{Acknowledgment}
This work is supported by the National Natural Science Foundation of China (Grant No. 61427811, U1836217, 61603391).
Special thanks to Ran He, Qi Li and Yuhao Zhu.


\small{
\bibliographystyle{aaai} \bibliography{egbib}

}
\end{document}